\documentclass[conference]{IEEEtran}
\IEEEoverridecommandlockouts
\usepackage{cite}
\usepackage{amsmath,amssymb,amsfonts}
\usepackage{graphicx}
\usepackage{textcomp}
\usepackage{xcolor}
\usepackage{flushend}
\usepackage[a4paper, left=0.51in, right=0.51in, top=0.75in, bottom=1.69in]{geometry}
\usepackage{epsfig}

\usepackage{support-caption}
\usepackage{subcaption}
\usepackage{calc}
\usepackage{amstext}
\usepackage{amsthm}
\usepackage{multicol}
\usepackage{pslatex}
\usepackage{multirow}

\usepackage{enumerate}
\usepackage{algorithm}
\usepackage[noend]{algpseudocode}

\newcommand{\etal}{et al. }
\newcommand{\eg}{e.g.,}

\def\BibTeX{{\rm B\kern-.05em{\sc i\kern-.025em b}\kern-.08em
    T\kern-.1667em\lower.7ex\hbox{E}\kern-.125emX}}
\IEEEoverridecommandlockouts
\IEEEpubid{%
    \makebox[\columnwidth]{%
        978-1-7281-7539-3/20/\$31.00~\copyright{}2020 IEEE\hfill%
    } \hspace{\columnsep}\makebox[\columnwidth]{ }%
}
\begin{document}

\title{Open-set Face Recognition for Small Galleries Using Siamese Networks}

\author{\IEEEauthorblockN{Gabriel Salomon\IEEEauthorrefmark{1}, 
Alceu Britto Jr.\IEEEauthorrefmark{2},
Rafael H. Vareto\IEEEauthorrefmark{3},
William R. Schwartz\IEEEauthorrefmark{3},
David Menotti\IEEEauthorrefmark{1}}
	\IEEEauthorblockA{\IEEEauthorrefmark{1}Vision, Robotics and Imaging Laboratory, Universidade Federal do Paran\'{a}, 82590300, Brazil}
	\IEEEauthorblockA{\IEEEauthorrefmark{2}PPGIA, Pontif\'{i}cia Universidade Cat\'{o}lica do Paran\'{a}, 80215901, Brazil}
	\IEEEauthorblockA{\IEEEauthorrefmark{3}Smart Sense Laboratory, Department of Computer Science, Universidade Federal de Minas Gerais, 31270901, Brazil}
	\textit{gsaniceto@inf.ufpr.br}}

\maketitle

\begin{abstract}
Face recognition has been one of the most relevant and explored fields of Biometrics. In real-world applications, face recognition methods usually must deal with scenarios where not all probe individuals were seen during the training phase (open-set scenarios). 
Therefore, open-set face recognition is a subject of increasing interest as it deals with identifying individuals in a space where not all faces are known in advance.
This is useful in several applications, such as access authentication, on which only a few individuals that have been previously enrolled in a gallery are allowed.
The present work introduces a novel approach towards open-set face recognition focusing on small galleries and in enrollment detection, not identity retrieval.
A Siamese Network architecture is proposed to learn a model to detect if a face probe is enrolled in the gallery based on a verification-like approach.
Promising results were achieved for small galleries on experiments carried out on Pubfig83, FRGCv1 and LFW datasets. State-of-the-art methods like HFCN and HPLS were outperformed on FRGCv1. 
Besides, a new evaluation protocol is introduced for experiments in small galleries on LFW.
\end{abstract}

\renewcommand\IEEEkeywordsname{Keywords}

\section{Introduction}
\label{sec:introduction}

Face recognition is a real-world necessity, mainly for human identification and surveillance. It presents many advantages over other biometrics (\eg  fingerprint and iris), as face images capture is non-intrusive and can be done at a distance~\cite{stan2011}.  
Moreover the availability of face samples on the Internet.

A common definition of face recognition assumes the existence of three tasks, as follows: verification, identification and watch list recognition~\cite{chellappa2010}.
Face verification is defined as a binary problem that consists of comparing two face images and then determining if both of them belong to the same individual.  
Closed-set face identification, or just identification, focus on determining to which gallery member a face image sample belongs.
Watch list recognition (or open-set face recognition), works similarly to the latter by also comparing the face sample to the gallery, but with two major distinctions:
i) the individual may not belong to the gallery;
ii) the face corresponding identity is only relevant if it was previously enrolled in the gallery.
More precisely, an open-set face recognition task addresses the problem of determining whether a face image is enrolled in a previously defined gallery and, in that case, retrieving the sample's corresponding enrolled identity.

\begin{figure}[!t]
\begin{center}
   \includegraphics[width=0.95\linewidth]{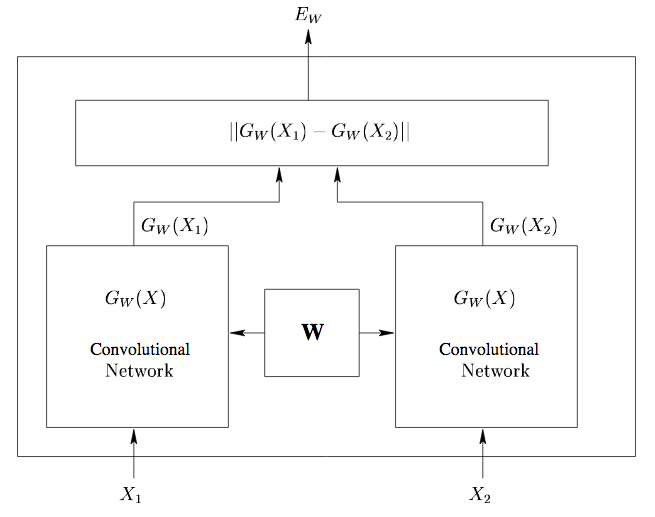}
\end{center}
\vspace{-12pt}
   \caption{Representation of a Siamese Network. It has two input images $X_1$, $X_2$ for each network (convolutional networks in this example). The architecture (or function $G$) is the same for both and the weights $W$ are shared. The outputs are joined at the end using Euclidean distance. 
   The final output is $E_W$. Source: Chopra \etal \cite{chopra2005learning}. }
\label{fig:siamesaex}
\vspace{-12pt}
\end{figure}

Machine learning methods usually learn a model that uses training data to map feature vectors to known classes~\cite{salomon2020deep, ruiz2020can}.
Testing these methods consists in dealing with unknown samples of known classes. 
However, dealing with unknown classes on testing is not trivial. 
Open-set face recognition methods have to deal with an ``unlimited'' number of unknown classes (subjects' identities) when compared to the countable quantity of individual identities enrolled in the gallery set~\cite{scheirer2013toward}.

It is important to stress that this work aims to determine whether a given face sample belongs to any individual registered in the gallery set, and not at retrieving its genuine identity.
In that direction, our proposal is directly comparable to works in the literature~\cite{vareto2017towards}.
Our main hypothesis is that Siamese Neural Networks (SNNs) can be effectively employed in open-set recognition.
A SNN is composed of two independent sub-networks with the same architecture and shared weights with their outputs joined at the end using a distance metric~\cite{bromley1994signature}.  
Each independent sub-network has its own input, and the whole network outputs a numeric distance value, indicating the similarity between the two inputs. Although the SNN performs face verification, the proposed architecture uses it to compute scores and perform open-set recognition.
Fig.~\ref{fig:siamesaex} illustrates the process.

Small galleries are the focus of the present work due to a scalability limitation of the present method on the experiments performed.
To the best of our knowledge, this is the first time that SNN are considered in open-set face recognition scenarios focusing on small galleries.
The term \textit{small gallery} refers to sets containing no more than 20 identities, but with no restrictions on the number of samples per identity. 
Many business establishments require visual authentication of small personnel, rarely exceeding fifteen or twenty employees.
The lack of sufficient classes and samples turns the small gallery recognition task into a difficult problem, especially due to overfitting.
Similarly, scarce face samples may reduce the models ability to learn general facial characteristics.


The proposed method can be divided into two stages: first, an SNN is trained on generated pairs of face samples that belong to the gallery. It is trained for the usual task (verification): learning to differentiate the members of the gallery.
In the second stage, the trained SNN is used to calculate the minimum Euclidean distance between a probe image and training samples from all members of the gallery. It is expected that the SNN learns a representation afterward expressed in terms of distance of a pair of images. These representations should allow establishing a separation between the probe images that belong to the gallery and the ones that do not.

Our main contribution is twofold. A new method using SNN for performing open-set recognition on small galleries; and a new experimental protocol for comparison of methods devoted to open-set small galleries. 
The method addresses a real necessity of algorithms that can perform well in constrained data (few examples and few identities known) and the protocol was defined to support the development of such methods.

Experiments performed on Experiment 4 of FRGCv1 dataset (Experiment 4)~\cite{phillips2005overview} achieved state-of-the-art results for small galleries in comparison with published work~\cite{vareto2017towards}.

\section{Related Work}
\label{sec:related}


A face verification-like approach towards open-set recognition is proposed in the current work, using CNN and SNN. In this respect, research regarding deep learning methods, open-set recognition and Siamese Networks for face recognition are reviewed.

One of the main focus of the face recognition research community in the last years has been the improvement of the feature extraction quality, which is measured by its ability to represent and differentiate face samples.
The most commonly used methods are based on Convolutional Neural Networks (CNNs). 

However, CNNs only became the most used feature extraction and learning techniques within the last decade.
Combined with GPU and large datasets, face recognition methods are capable of surpassing human performance using deep network architectures, achieving state-of-the-art performance on many surveillance and biometrics tasks~\cite{taigman2014deepface, parkhi2015deep, schroff2015facenet, sun2015deepid3, abdalmageed2016face, vareto2017verification}.

One of the most well-established networks with a focus on face recognition is the VGGFace~\cite{parkhi2015deep}, introduced in 2015. 
VGGFace is pre-trained on 2.6 million face images and achieves 98.95\% accuracy on the Labeled Faces in the Wild (LFW) dataset~\cite{huang2008labeled} under the face verification task. 
Conversely to most works in the literature, we design an open-set face recognition approach following a face-verification strategy.

According to Kemelmacher~\etal \cite{kemelmacher2016megaface}, recent identification and verification face recognition approaches are not very promising when the gallery set is upscaled, that is, when the set of known individuals escalates towards thousands or millions of subjects.
For that reason, scalability is drawing the attention of several researchers~\cite{pinto2011scaling, de2015classification, taigman2015web, abdalmageed2016face}.
Vareto~\etal \cite{vareto2017towards} concentrate on the open-set face recognition problem as they assess both Partial Least Squares (PLS) and Multi-Layer Perceptron (MLP) classification models in pursuance of an algorithm that is not directly dependent on the gallery set size.
In fact, the authors build a collection of either PLS or MLP binary models, defined as hashing functions, and a voting system scheme (candidate list) to determine whether the queried subject is known or unknown.
If a subject stands out from the other individuals of the candidate list, the subject is recognized; otherwise, the subject is discarded since the method considers s/he is not enrolled in the gallery set.

Siamese Networks are used for metric learning and were originally developed for verifying signatures using Neural Networks~\cite{bromley1994signature}.
A few years later, they were employed in face verification for the first time, using a new loss function based on Energy-Based Models (EBM), called Contrastive Loss~\cite{chopra2005learning}. 
Since the boom of CNNs in face recognition, Siamese Networks have been widely used for face verification in large datasets, obtaining great performance over other methods~\cite{parkhi2015deep, taigman2014deepface, yang2017neural}.
As Siamese Networks are a metric learning method, they make use of similarity scores to perform recognition.
Several works have also proposed approaches based on thresholding similarity-like scores, but focusing generally on open-set face recognition~\cite{li2005open, pinto2008real, dos2014extending, taigman2015web,sun2015deepid3, abdalmageed2016face, gunther2017toward, liu2017sphe}.

\section{Proposed Method}
\label{sec:method}
\begin{figure*}[ht!]
\begin{center}
   \includegraphics[width=0.9\linewidth]{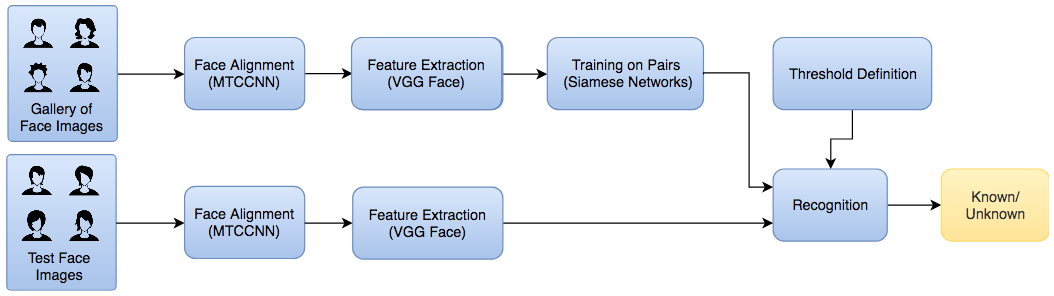}
\end{center}
   \caption{Pipeline for proposed method. The gallery is used for training the Siamese Networks, after face alignment/crop and feature extraction. At the end, a threshold is defined. The test face images are aligned/crop and their features are extracted to perform recognition. The recognition stage assigns a binary label to each test sample: ``Known" or ``Unknown".}
\label{fig:mymethod}
\vspace{-12pt}
\end{figure*}

The proposed method consists of 4 steps: i) face alignment; ii) feature extraction; iii) training on pairs; iv) recognition.
The two first steps are straightforward approaches.
On the third step, negative and positive pairs are generated from the face gallery samples to train the SNN to learn how to differentiate between the gallery members.
Using the trained SNN, in the last step, a test sample is compared to selected training samples from each gallery member and the recognition is performed based on the similarity score.
Each stage is detailed in the next sections and illustrated in Fig. \ref{fig:mymethod}.

\subsection{Face Alignment}

Alignment is conducted in all face images using Joint Face Detection and Alignment using Multi-task Cascaded CNNs~\cite{zhang2016joint}.
The Multi-task Cascaded CNN can perform both face detection and alignment at the same time. 
There are three stages to perform both tasks:
\begin{enumerate}[i.]
    \item
a shallow CNN produces a large set of windows in the image that might be a face or not;
    \item
a more complex CNN reject most of the windows, which are unlikely to contain faces;
   \item
an even more complex CNN analyzes the results refining it and outputs landmarks positions for aligned faces.
\end{enumerate}
This state-of-the-art method was used in the pre-processing stage, as face cropping and alignment may improve feature extraction discrimination capability. 
\subsection{Feature Extraction}
The extraction of features from the images for training and testing was made using a Deep
Convolutional Network: VGG Face \cite{parkhi2015deep}. 
The network has 21 layers (18 for convolution and pooling and 3 for classification, respectively) plus a Softmax function at the end for prediction purposes. 
VGG is pre-trained on a celebrity database that contained 2.6 million images of 2,622 individual identities.
Aiming only feature extraction, the Softmax function is removed, and the resulting feature vector for every image has 2,622 dimensions.

\subsection{Siamese Networks}
\begin{figure}[ht]
   \includegraphics[width=0.8\linewidth]{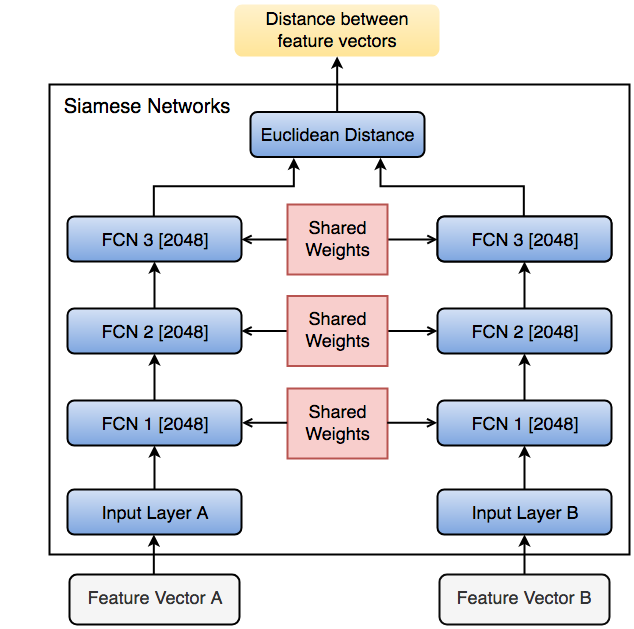}
   \caption{SNN architecture. The weights of each FCN are shared among the two separated internal pipelines of the network, this means that the weights are the same (not that they are connected). In the end, the output is the euclidean distance.}
\label{fig:siamese}
\vspace{-14pt}
\end{figure}

The SNN architecture is composed of two identical 3-layer sub-networks, each layer is fully connected and composed of 2,048 neurons.   
The architecture and the number of neurons were obtained empirically. 
The input layer has size 2,622, which is the exact size of the output of the VGG Face feature extractor.
Fig.~\ref{fig:siamese} presents the network architecture.

\subsubsection{Pair Generation}
The maximum possible number of training samples pairings is the combination two at a time of the $n$ samples, i.e., $\binom{\ n \ }{2} = \mathcal{O}(n^2)$.
This can be costly, considering that few iterations on training over almost $n^2$ pairs can take very long, as neural networks are expensive to train. 
Subsampling is needed to reduce the number of pairs used in training. 

Also, an evaluation on how the pairs choice impacts the performance is desirable.
Therefore, two methods of pair subsampling were evaluated.
These specific techniques were selected empirically on initial experiments.

\begin{figure*}[!ht]
   \begin{subfigure}{.5\textwidth}
      \centering
      \includegraphics[width=.65\linewidth]{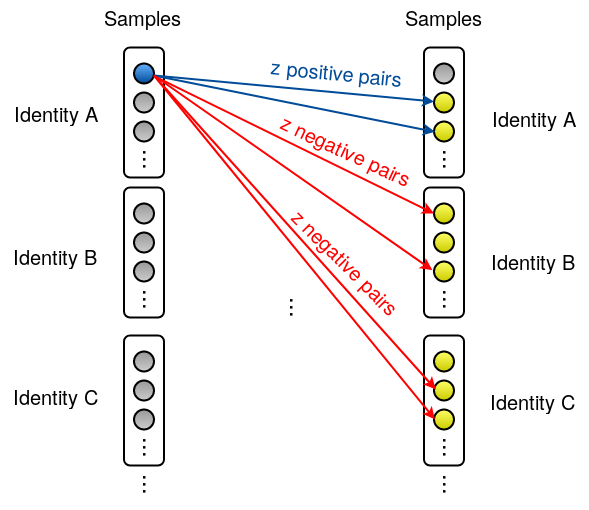}  
      \caption{Pairing Algorithm 1.}
      \label{fig:sub-first}
    \end{subfigure}
     \begin{subfigure}{.5\textwidth}
      \centering
      \includegraphics[width=.63\linewidth]{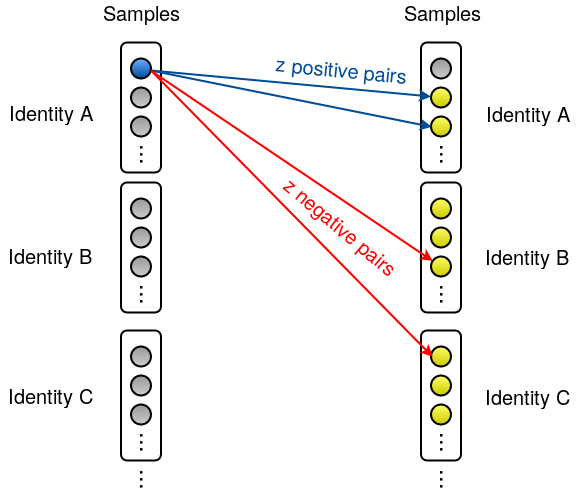}  
      \caption{Pairing Algorithm 2.}
      \label{fig:sub-second}
    \end{subfigure}
    \caption{Comparison of P1 and P2. All ``Samples" columns represent the same samples.
    Note that in P1, the $z$ pairs are formed between the same identity and $z$ pairs for each one of the other identities. P2, in contrast, generate $z$ pairs choosing randomly samples for any other identity (generating less pairs).
    }
\vspace{-10pt}
\end{figure*}

For explanation purposes, they are denominated: Algorithms P1 and P2.
The training samples set is denoted as $\mathcal{S}$, and the set of identities $\mathcal{I}$.
The function $id()$ returns the identity associated with the sample supplied as parameter. 
The procedure $rand\_sample()$ receives a set of identities and returns a random sample that belongs to one of the identities provided.
In the end, each algorithm return two sets of pairs, $PP$ and $PN$, which correspond to the set of pairs of the samples from the same (positive) and different persons(negative), respectively. 
Algorithm P1 is described below and illustrated in Fig.~\ref{fig:sub-first}.
\begin{algorithm}
\caption{Pairing 1}
\label{alg:p1}
\begin{algorithmic}[1]
\State $PN = [\ ]$
\State $PP = [\ ]$

\For{$x$ in $S$}
\For{$i$ in $I-id(x)$}
\State $c = 0$
\While{$c<z$}  
\State $PN+=[x, rand\_sample(i)]$
\State $PP+=[x, rand\_sample(id(x))]$
\State $c+=1$
\EndWhile
\EndFor
\EndFor
\end{algorithmic}
\end{algorithm}

Both algorithms generate equal numbers of positives and negatives pairs, for balancing purposes.
\textbf{P1} generates pairs using the individual samples \textit{versus} every other subject in the gallery. 
For each training sample, pairings with $z << n$ random samples from every other individual in the gallery are generated. 
In cases where $z$ is greater than the number of samples of each individual, repeated pairs will occur.
On the other hand, \textbf{P2} generates fewer pairs, only pairing samples with a few other individuals randomly chosen. 
Every training sample is paired with $z$ random samples from other random identities.
Algorithm P2 is presented below and illustrated in Fig.~\ref{fig:sub-second}.

\begin{algorithm}
\caption{Pairing 2}\label{alg:p2}
\begin{algorithmic}[1]
\State $PN = [\ ]$
\State $PP = [\ ]$
\For{$x$ in $S$}
\State $c = 0$

\While{$c<z$}  
\State $PN+=[x, rand\_sample(I-id(x))]$
\State $PP+=[x, rand\_sample(id(x))]$

\State $c+=1$
\EndWhile
\EndFor

\end{algorithmic}
\end{algorithm}

Observe that for P1, $n \times 2(k-1) \times z$ pairs are generated, where $k=|\mathcal{I}|$, while for P2 only $n \times 2 \times z$ pairs are generated and $2$ stands for the ``positive'' and ``negative'' sets.

\subsubsection{Loss and Distance Functions}

The loss function chosen is the Contrastive Loss~\cite{chopra2005learning,hadsell2006dimensionality}. 
Contrastive Loss takes pairs of samples and maps them into a space where intra-class samples reside in neighbor areas and inter-class samples are far away.

The weights ($W$) of the network are the parameters of the mapping function (called $G$), that reduces the dimensionality by mapping two feature vectors $\vec{X}_{1}$ and $\vec{X}_{2}$ into $G_W(\vec{X}_{1})$ and $G_W(\vec{X}_{2})$ and computing the distance measure $D_W(\vec{X}_{1},\vec{X}_{2})$, i.e., $D_W(\vec{X}_{1},\vec{X}_{2}) = || G_W(\vec{X}_{1}) - G_W(\vec{X}_{2}) ||$.
%
%
%
For simplicity $D_W(\vec{X}_{1},\vec{X}_{2})$ is denoted $D_W$ and the final loss function~\cite{chopra2005learning} is defined as
\begin{equation}
\resizebox{0.85\hsize}{!}{
$ L(W,Y,\vec{X}_{1},\vec{X}_{2}) =  \frac{(1-Y)(D_W)^2 + Y\{\max(0,m-D_W)\}^2}{2}$}
\end{equation}
\noindent where $Y$ is 0 if the feature vectors belong to the same person and 1 if they represent different identities, and the variable $m$ is a predefined margin around $G_W(\vec{X}_{1})$ or $G_W(\vec{X}_{2})$ that should be greater than 0.
In other words, if  $\vec{X}_{1}$ and $\vec{X}_{2}$ belong to different individuals and $D_W(\vec{X}_{1},\vec{X}_{2})$ is smaller than the $m$ margin, the loss function improves their separation, otherwise, no margin changes are made.

The Siamese Network's distance chosen is the euclidean distance. 
Other commonly used distances are cosine similarity and Manhattan distance, but in the experiments performed with Contrastive Loss, euclidean distance outperformed both.

\section{Experiments}
\label{sec:experiments}

In this section, the experiments executed to validate the proposed approach are described.
For such, the evaluation metrics, datasets employed and the experimental protocols are described.
For literature comparison, we chose the work of Vareto \etal\cite{vareto2017towards}, taking into account the clarity on the evaluation protocol which assists reproducibility of experiments, enabling a fair comparison.

\subsection{Metrics}

AUC (Area Under Curve) from the ROC (Receiver Operating Characteristic) curve is the metric selected for method evaluation.
It is a well-known technique as it describes the algorithm performance on various thresholds.
%
ROC curve shows the relation between TPR (True Positive Rate) and FPR (False Positive Rate).
ROC curves with AUC closer to $1$ indicate better performance, while the ones with AUC close to $0.5$ indicate poor performance. Values below $0.5$ are worst than random guessing.



\subsection{Datasets}

One of the most challenging and used datasets for Face Recognition is certainly LFW\cite{huang2008labeled}. However, as previously stated, there is no universal protocol for open-set evaluation and for literature comparison.
The first two datasets described below are used for literature comparison, and, the third one is used for evaluating the proposed (new) protocol.

\noindent \textbf{\textit{PubFig83}}: 
Pubfig83 is a subset of the original PubFig \cite{kumar2009attribute}. It has only 83 subjects and 13,838 uncontrolled images (average of 166 samples per subject). Ranges from 89 to 367 samples available per subject.

\noindent \textbf{\textit{FRGCv1}}:
This is the dataset provided for the Face Recognition Grand Challenge v1.0  \cite{phillips2005overview}. It will be used to compare with Vareto \etal experiments.
It has a total of 50,000 images and 6 experiments. Tests were only conducted on experiment 1 and 4. The first experiment is composed of 760 controlled images and the fourth has 152 controlled photos + 1064 uncontrolled ones. Both experiments have 152 distinct subjects.

\noindent \textbf{\textit{LFW}}: 
LFW is a well-known dataset among face recognition researchers, in fact, one of the most used for face recognition. It consists of 13,233 uncontrolled images of 5,749 people. Many subjects in this dataset have only 1 sample available. Some have more than 15 images.

\subsection{Experimental Protocols}
The main issue in choosing the experimental protocol (EP) lies upon the lack of consensus among authors on which is the best protocol for open-set recognition evaluation. Also, the protocol is usually not clearly stated making reported results unable to be reproduced and compared fairly.

As we are comparing our proposal to the one by Vareto \etal \cite{vareto2017towards}, and their protocol is well described, we used their protocol, which we refer to as \textit{Experimental Protocol I}.

We also propose a slight modification on EP-I focusing on \textit{small galleries}, thus generating the so-called \textit{Experimental Protocol II}.
The difference between these two protocols is that in the EP-I, we choose a \textit{percentage} of subjects, i.e., 10\%, 50\% and 90\% of identities in the dataset to be part of the known individual, while in the EP-II an \textit{absolute} number of subjects (e.g., 5, 10, 15 and 20) is chosen.

In both protocols, the database is split into training and testing data. 
The training data has only known individuals (individuals that should be recognized by the method). The testing data has known individuals unseen samples (samples of the individuals that the method has never seen before) and unknown individuals (samples of people the method has never seen before). 

The whole dataset is first divided between known/unknown individuals. 
The known subjects are evaluated in different scenarios, i.e.,  10\%, 50\% and 90\% of all dataset identities for EP-I and 5, 10, 15, and 20 subjects for EP-II in the dataset and the remainder is considered as unknown and is only used for~testing.

For each known identity, the samples are split in half: 50\% for training and 50\% for testing.
Each experiment on each database is repeated 10 times by randomly splitting the data as aforementioned, executing the method and, in the end, plotting the ROC curves with AUC.
The mean AUC is computed, along with the standard deviation.

\subsection{Experimental Results}

In this section, we report results yielded by our proposed method using:
i) the EP-I (proposed by Vareto \etal \cite{vareto2017towards}) on the Pubfig83 \cite{pinto2011scaling} and FRGCv1 \cite{phillips2005overview} (exp. 1 and 4) datasets (Table~\ref{tab:results});
ii) The new proposed EP-II on the LFW dataset \cite{huang2008labeled} (Table~\ref{tab:lfw});
iii) the EP-I on the FRGCv1 dataset - exp. 4 for comparison with the literature (Table~\ref{tab:cmpfrgc}).

\begin{table}[!tb]
\footnotesize
    \centering
    \caption{Results for experiments on Pubfig83 and FRGCv1 (Experiments 1 \& 4) dataset using EP-I.}
    \label{tab:results}
    \resizebox{0.95\columnwidth}{!}{%
    \begin{tabular}{|c|c|c|c|c|}
         \hline
         \multirow{3}{*}{Pair} 
              &       & \multicolumn{3}{c|}{Datasets} \\
         \cline{3-5}
              & Known & \multirow{2}{*}{Pubfig83}
                                  & FRGCv1 & FRGCv1 \\
              & Rate  &           & Exp. 1 &  Exp. 4 \\
                       \cline{3-5} 
              &       & AUC & AUC  & AUC \\
        
         \hline         
         P1   & \multirow{2}{*}{10\%} & $.981 \pm .008$ 
                                      & $.996 \pm .002$  
                                      & $.938 \pm .027$ \\ 
         P2   &                       & $.981 \pm .007$ 
                                      & $.996 \pm .003$ 
                                      & $.904 \pm .032$  \\
         \hline
         P1   & \multirow{2}{*}{50\%} & $.922 \pm .015$ 
                                      & $.986 \pm .004$ 
                                      & $.868 \pm .013$  \\
         P2   &                       & $.916 \pm .018$ 
                                      & $.986 \pm .003$ 
                                      & $.816 \pm .019$  \\
         \hline
         P1   & \multirow{2}{*}{90\%} & $.856 \pm .017$ 
                                      & $.972 \pm .018$ 
                                      & $.797 \pm .030$  \\
         P2   &                       & $.854 \pm .025$  
                                      & $.974 \pm .008$  
                                      & $.770 \pm .021$  \\
         \hline
    \end{tabular}
    }
    \vspace{-5pt}
\end{table}

\noindent \textbf{\textit{PubFig83 (EP-I)}}: 
The proposed method achieved great results in Pubfig83 for small galleries (few subjects known). 
Although with the increase of the number of subjects, the AUC dropped significantly in 90\% known. 
Statistically, there is no difference between the pairings P1 and P2.

\noindent \textbf{\textit{FRGCv1 - Experiment 1 (EP-I)}}: 
In experiment 1 from FRGCv1 dataset outstanding results are obtained as well.
We believe that this is due to the fact that there are only controlled images on this dataset. 
There is no great challenge in discriminating controlled images. 
There is no difference in the results concerning the type of pairing used in training. 
The increase in gallery size did not influence dramatically the AUC, the difference is not statistically significant.

\noindent \textbf{\textit{FRGCv1 - Experiment 4 (EP-I)}}:
When using small size gallery better results than the bigger galleries are obtained, as in other experiments.
There is a significant discrepancy in the AUC rate for a small gallery between P1 and P2, where P1 shows better performance.

\noindent \textbf{\textit{LFW (EP-II)}}: 
%
The method performed well for small galleries, taking into account that LFW is unbalanced in the number of samples per subject and uncontrolled. 
There is no discrepancy in the AUC rates for small galleries between P1 and P2.

\begin{table}[!ht]
    \centering
    \caption{Results of the experiments on LFW using EP-II.}
    \label{tab:lfw}
    \resizebox{0.53\columnwidth}{!}{%

    \begin{tabular}{|c|c|c|}
         \hline
         Pair & Known & ROC (AUC)  \\
         \hline         
         P1   & \multirow{2}{*}{5} & $0.971 \pm 0.019$   \\
         P2   &                     & $0.969 \pm 0.019$   \\
         \hline 
         P1   & \multirow{2}{*}{10} & $0.967 \pm 0.028$   \\
         P2   &                      & $0.967 \pm 0.030$   \\
         \hline
         P1   & \multirow{2}{*}{15} & $0.976 \pm 0.011$   \\
         P2   & &  $0.946 \pm 0.006$    \\
         \hline
         P1   & \multirow{2}{*}{20} & $0.972 \pm 0.007$   \\
         P2   & & $0.977 \pm 0.008$   \\
         \hline
    \end{tabular}
    }
        \vspace{-5pt}
\end{table}

\noindent \textbf{\textit{Literature Comparison}}: 
As shown in Table \ref{tab:cmpfrgc}, for a small number of identities (small galleries) the proposed method with SNN can surpass state-of-the-art methods in experiment 4 on FRGCv1. 
Note that we report the best result obtained by P1 ($.938 \pm .027$) -- from Table~\ref{tab:results}. 
As there are approximately 150 identities, there are 15 subjects in the gallery. 
When there are 50\% known people in the gallery (75 identities), no statistical difference between the methods occurs.

\begin{table}[!ht]
	\centering
    \caption{Comparison with the work of Vareto \etal on FRGCv1 - experiment 4.}
    \label{tab:cmpfrgc}
     \resizebox{0.83\columnwidth}{!}{%
    \begin{tabular}{|c|c|c|c|c|}
    \hline
      \multicolumn{2}{|c|}{\multirow{2}{*}{Known Individuals}} & \multicolumn{3}{c|}{AUC} \\ \cline{3-5}
      \multicolumn{2}{|c|}{}
               & 10\% & 50\% & 90\% \\
     \hline 
    \multirow{2}{*}{SN (proposed)} 
        & mean &  0.938  &  0.868  &  0.783 \\ 
        & std  & $0.027$ & $0.013$ & $0.029$
    \\ \hline
    \multirow{2}{*}{HPLS \cite{vareto2017towards}} 
        & mean &  0.794  &  0.850  &  0.856 \\
        & std  & $0.078$ & $0.009$ & $0.022$
    \\ \hline
    \multirow{2}{*}{HFCN \cite{vareto2017towards}}
        & mean &  0.872  &  0.877  &  0.856 \\
        & std  & $0.015$ & $0.022$ & $0.014$
    \\ \hline
    \end{tabular}
     }
    \vspace{-5pt}
\end{table}

\section{Conclusions}
\label{sec:conclusion}

We introduced a recognition system, with well-defined pipelines, that employs a SNN for open-set face recognition, a task that still has not been extensively researched. We showed that SNNs outperform state-of-the-art methods like HPLS and HFCN on small galleries, being useful in applications on which reliability is more relevant than scalability.
Another contribution is the Experimental Protocol II proposed for small galleries. 

For future work, our goals are improving pairing selection, testing new loss functions and also combining SNNs with other metric learning methods to enhance performance.




\section*{Acknowledgments}

The authors would like to thank the National Council for Scientific and Technological Development -- CNPq (Grants \#438629/2018-3, \#309953/2019-7, \#313423/2017-2,  \#311053/2016-5, and \#428333/2016-8), the Minas Gerais Research Foundation -- FAPEMIG (Grants APQ-00567-14 and PPM-00540-17), the Coordination for the Improvement of Higher Education Personnel -- CAPES (DeepEyes Project), and also acknowledge the support of NVIDIA Corporation with the donation of the Titan Xp GPU used for this research.

\bibliographystyle{ieeetr}
\bibliography{example}

\end{document}